\documentclass{article}
\usepackage[preprint,dblblindworkshop]{neurips_2026}
\workshoptitle{Agentic AI for Biological Discovery (AgenticLS)}
\usepackage[utf8]{inputenc}
\usepackage[T1]{fontenc}
\usepackage{hyperref}
\usepackage{url}
\usepackage{booktabs}
\usepackage{array,longtable}
\usepackage{amsfonts}
\usepackage{microtype}
\usepackage{graphicx}
\usepackage{amsmath}
\usepackage{xcolor}
\usepackage{listings}
\hypersetup{hidelinks}
\title{Does reflective LLM feedback survive\\a wet-lab-scale label budget?}
\author{%
  Haoxuan Zeng \qquad Xin Luo\thanks{Corresponding author: \texttt{luosanj@umich.edu}.}\\
  University of Michigan, Ann Arbor, Michigan, USA\\
  \texttt{zchx@umich.edu} \qquad \texttt{luosanj@umich.edu}
}
\begin{document}
\maketitle
\begin{abstract}
Can reflective language-model feedback improve a biological predictor when each cycle buys
only one plate of labels? We separate three possible sources of benefit: proposing model edits,
choosing measurements, and revising decisions through repeated feedback. We replay $36$
ProteinGym assays, starting with $96$ labels and acquiring $96$ per round for six rounds,
with a sealed test set and a shared edit schema. The pre-specified comparison of full LLM
control against random edits detects no advantage in test rank correlation
($\Delta z=-0.008$, $95\%$ CI $[-0.034,+0.018]$); equivalence is also unestablished.
Exploratory ablations nevertheless identify useful model editing. With random acquisition fixed,
LLM edits improve Fisher-transformed test Spearman by $0.043$ over no edits; with acquisition
fixed to upper confidence bound, they improve normalized top-variant recovery by $0.023$.
LLM acquisition alone does not detectably outperform fixed adaptive rules, and repeated
feedback does not detectably outperform one initial recommendation under full control.
Thus, the observed gains support constrained specialist editing, while the incremental value
of recursive reflection remains unestablished at this label budget.
\end{abstract}
\section{Introduction}
\label{sec:intro}
An agent for biological discovery must decide both how to improve its predictor and what to
measure next. A general-purpose large language model (LLM) could coordinate these decisions by
reading a specialist's diagnostics, proposing changes, and revising them after new measurements.
In a small experimental campaign, however, the same scarce labels must support training and
evaluation. As acquisition changes which variants are labeled, it also changes the evidence
used to judge the next model edit. A better final predictor therefore need not imply that
repeated reflection was responsible for the improvement.

We examine this distinction in protein-fitness prediction with batches of $96$ labels, matching
the nominal capacity of one assay plate. The LLM can edit a supervised predictor's data
processing, features, training recipe, and acquisition rule. Only the specialist is refitted:
the generalist and protein encoder remain frozen. This setting tests a component of
feedback-driven specialist improvement, rather than an agent that updates its own weights.
We ask whether LLM proposals help, whether control of acquisition adds value, and whether
repeated feedback improves on a single recommendation.

Across $36$ assays, full control shows no detected advantage over random edits on the pre-specified
ranking endpoint. Exploratory comparisons locate positive results in a narrower role: model
editing under a prescribed acquisition rule. The benefit depends on the objective---global
ranking or recovery of high-fitness variants---and does not establish an incremental benefit
from repeated feedback. This decomposition provides a practical way to evaluate claims of
biological recursive self-improvement without conflating proposal quality, acquisition, and reflection.

\section{Related Work}
\label{sec:related}
\paragraph{Biological prediction and experiment selection.}
Combining evolutionary information with measured fitness supports learning from limited
assay data \citep{hsu2022learning}. EVOLVEpro couples protein-language-model features with
iterative experimental selection \citep{jiang2025evolvepro}, while BioDiscoveryAgent uses
LLM reasoning to propose genetic perturbations \citep{roohani2025biodiscoveryagent}.
Outside biology, label-efficient acquisition has been studied for optical-flow training
\citep{yuan2022opticalflow} and industrial multimodal content understanding, where kNN-based
Latent Space Broadening guides sample selection \citep{sun2025audioenhanced}.
We evaluate the added value of generalist-guided specialist editing against fixed acquisition
rules and non-LLM editing controls.

Model-based diagnosis has also been studied in biological workflows:
Li uses experimental outcomes and records of reagents and technicians
to identify sources of failure \citep{li2022root}.
Our study instead examines whether diagnostic feedback improves
protein-fitness prediction under a restricted assay-label budget.

\paragraph{Feedback-driven improvement.}
OPRO uses previous candidates and scores to guide optimization \citep{yang2024large}; GEPA
uses textual feedback to evolve prompts \citep{agrawal2026gepa}. The Darwin G\"odel Machine
extends self-improvement to an agent's code \citep{zhang2026darwin}.
In a different domain, constrained integration embeds foundation-model predictions in
discrete-choice models while preserving economic structure
\citep{wang2026embeddingfoundationmodelpredictions}.
Here, the editable object is a supervised biological predictor with a bounded configuration
space. Its evaluation signal comes from the same purchased labels used for training, and
adaptive acquisition changes their distribution. We therefore evaluate repeated feedback
separately from the ability to propose a useful initial configuration.

\paragraph{Resource-constrained agents and biomedical data processing.}
CE-Agent uses dynamic cloud--edge offloading and trajectory distillation to balance
agent capability with latency and communication costs \citep{huang2026agent}.
In clinical machine learning, \citet{liang2025revealing} use an LLM to extract treatment
adherence from clinical notes and study how non-adherence affects causal estimates
and predictive models. These applications motivate separating useful LLM-assisted
operations from the added value of repeated reflection. Our study focuses on the
cost of acquiring assay labels that both train the predictor and evaluate proposed edits.

\section{Experimental Setup}
\label{sec:setup}
\paragraph{Assays, features, and label budget.}
We replay existing single-substitution measurements from ProteinGym \citep{notin2023proteingym}.
A filter fixed before the main results selects $36$ assays and $101{,}201$ variants:
$1{,}500$--$6{,}000$ single mutants, protein length $\leq400$, and one assay per UniProt accession.
Features comprise frozen ESM-2 150M mean embeddings \citep{lin2023esm2}, a masked-marginal
zero-shot score \citep{meier2021esm1v}, and optional substitution one-hot encodings.
Each run starts with $96$ labels and acquires $96$ in each of six rounds, ending at $672$.
The primary split seals $25\%$ of variants for testing; a harder split instead holds out
$25\%$ of residue positions. Test labels and scores never enter controller feedback.
This is an offline replay, with no new wet-lab measurements.

\paragraph{Controller and acceptance rule.}
Five-fold cross-validation (CV) on purchased labels supplies the only selection signal.
The full controller receives CV Spearman and a bootstrap interval, stratified errors,
uncertainty calibration, the current configuration, and edit history. It proposes changes
to $15$ fields covering curation, pseudo-labeling, features, training, and acquisition.
A shared validator rejects unknown fields and illegal categories. Non-acquisition edits are
kept only if CV Spearman improves; acquisition edits bypass this gate for every arm because
CV on existing labels cannot evaluate an unmeasured batch. The main controller uses the
configured Opus~5 identifier without external tools; identifiers and prompts are detailed in
Appendix~\ref{app:protocol}.

\paragraph{Controls and ablations.}
\emph{No edits} refits the default ridge predictor with random acquisition.
\emph{Random edits} matches a frozen pilot's distribution of LLM edit sizes; \emph{local search}
uses inner CV with nominal $1\times$ or $5\times$ LLM-call wall time, the latter capped at $24$
evaluations. A static augmented ridge recipe supplies another non-LLM baseline.
To separate model editing from measurement selection, we fix acquisition to random sampling
or upper confidence bound (UCB), or let the LLM edit acquisition alone. Fixed greedy, UCB,
and diversity-greedy rules provide acquisition controls. To test feedback, we compare the full
controller with scalar-only diagnostics and with one recommendation in round~1, followed by
retraining under its gated configuration. Additional controls appear in the appendix.

\paragraph{Endpoints and inference.}
The primary endpoint averages sealed-test Fisher $z=\operatorname{arctanh}(\rho)$, where $\rho$
is Spearman correlation, over rounds 3--6. The secondary endpoint averages normalized
top-variant recovery, $R_{96}=|P_{96}\cap T_m|/\min(96,m)$, over the same rounds: $P_{96}$
contains the $96$ highest predictions and $T_m$ the true top $m=\operatorname{round}(0.1n_{\rm test})$
variants. This equals recall when $m\leq96$ and precision when $m>96$. Both endpoints score unseen
test variants, not the acquired batch.
We average seeds $0$--$2$ within each assay, then form paired $t$ intervals across $36$ assays.
Splits and random streams are shared across arms; adaptive policies can still acquire different
labels. Full LLM versus random edits on the random split is the pre-specified primary contrast,
with a $0.02$ equivalence margin in $z$. Other comparisons are exploratory, with
Benjamini--Hochberg (BH) correction within each split, seed set, and endpoint family.
The random split was selected after a four-assay pilot, search was redesigned after partial
results, and an earlier grid with incorrectly keyed random streams was discarded.

\section{Results}
\label{sec:results}
\begin{figure}[t]
\centering
\includegraphics[width=\textwidth]{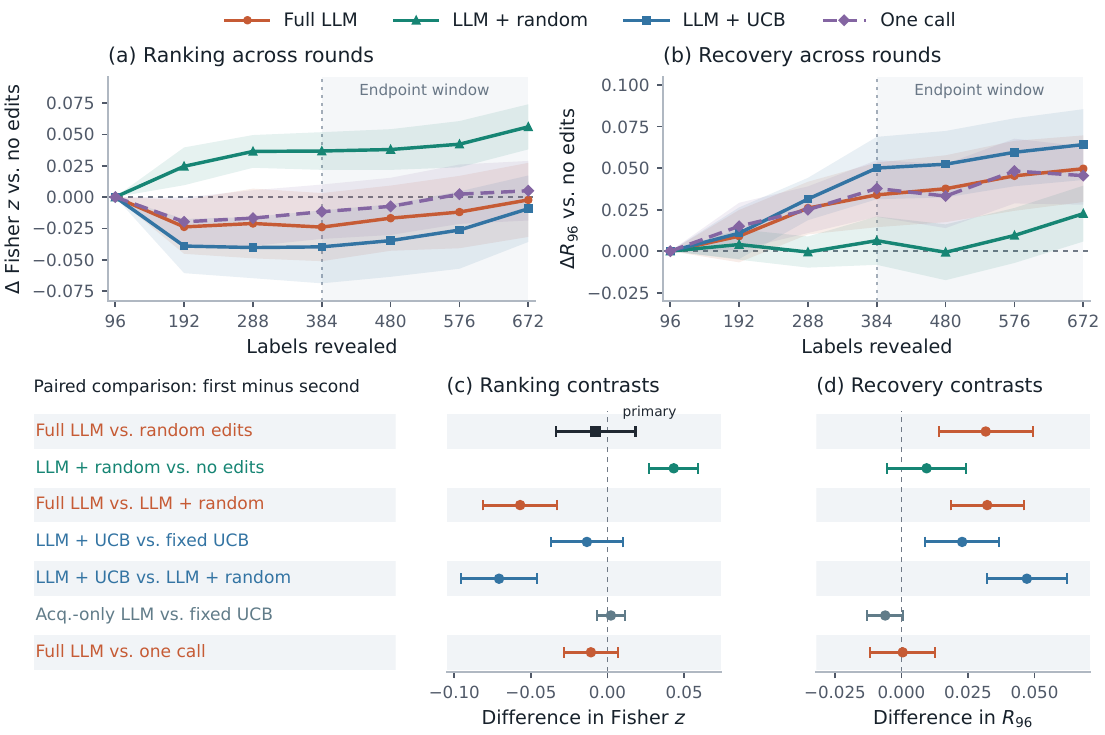}
\caption{How model editing, acquisition, and feedback affect the two objectives.
\textbf{(a,b)} Improvement over no edits as labels accumulate; shading gives pointwise $95\%$
intervals across $36$ assays, with seeds $0$--$2$ averaged within assay. The gray window marks
rounds 3--6 used for endpoint analysis. \textbf{(c,d)} Paired endpoint differences and $95\%$
$t$ intervals; positive values favor the first arm. Only the marked ranking comparison is
confirmatory. LLM + random/UCB fixes acquisition while allowing model edits.
} 
\label{fig:main}
\end{figure}
\begin{table}[t]
\centering
\caption{Random split, $108$ runs per arm, averaged over rounds 3--6. Acquisition is prescribed
unless marked editable or chosen once. $R_{96}$ is normalized top-variant recovery;
Appendix~\ref{app:controls} reports all $20$ complete arms.}
\label{tab:main}
\small
\begin{tabular}{@{}llrr@{}}
\toprule
Controller & Acquisition & Mean $z$ & $R_{96}$ \\
\midrule
No edits & Random & 0.603 & 0.349 \\
Random edits & Editable & 0.597 & 0.359 \\
Local search ($5\times$) & Editable & 0.623 & 0.354 \\
Full LLM & Editable & 0.590 & 0.391 \\
LLM, one call & Chosen once & 0.600 & 0.390 \\
\midrule
LLM model edits & Random & 0.647 & 0.358 \\
\midrule
No edits & UCB & 0.589 & 0.383 \\
LLM model edits & UCB & 0.576 & 0.406 \\
LLM acquisition only & Editable & 0.591 & 0.377 \\
\bottomrule
\end{tabular}

\end{table}

\paragraph{Full control shows no detected gain on the primary ranking endpoint.}
Full LLM minus random edits is $-0.008$ Fisher $z$ ($95\%$ CI $[-0.034,+0.018]$, $p=0.55$;
$20/36$ assay wins). Equivalence is also unestablished at the pre-specified margin
(two one-sided tests, $p=0.17$). Thus, neither superiority nor equivalence is established.
The secondary endpoint differs: $R_{96}$ increases by $0.032$ ($[0.014,0.049]$, $q=0.003$).
This divergence motivates examining which decisions help each objective (Figure~\ref{fig:main}).

\paragraph{Model edits help when acquisition is prescribed.}
With random acquisition fixed, LLM model edits gain $0.043$ in $z$ over no edits
($[0.027,0.059]$, $q<10^{-3}$; $31/36$ wins) and $0.024$ over nominal $5\times$ search
($[0.007,0.040]$, $q=0.019$). They also exceed full control by $0.057$
($[0.033,0.081]$). This ranking gain does not translate into a detected recovery gain over
no edits: $\Delta R_{96}=0.009$ ($[-0.005,+0.024]$).
Under fixed UCB, the pattern reverses: model editing improves $R_{96}$ by $0.023$ over UCB
alone ($[0.009,0.037]$, $q=0.005$), while its ranking difference is inconclusive
($\Delta z=-0.013$, $[-0.037,+0.010]$). Prescribed acquisition therefore enables useful
editing, but the benefit is specific to the evaluation objective.

\paragraph{LLM acquisition alone adds no detected benefit over fixed rules.}
Fixed adaptive policies achieve $R_{96}=0.380$--$0.384$, compared with $0.349$ for random
acquisition without edits. Acquisition-only LLM control reaches $0.377$, with no detected
advantage over fixed UCB ($\Delta R_{96}=-0.006$, $[-0.013,+0.001]$, $q=0.114$).
Moreover, switching the model-editing arm from random acquisition to UCB increases recovery
by $0.047$ ($[0.032,0.062]$) while reducing ranking by $0.071$ ($[0.046,0.095]$).
The full loop's recovery improvement therefore cannot be taken as evidence of better global
ranking, or attributed to a uniquely effective LLM acquisition policy.

\paragraph{The separation persists across acquisition rounds.}
Fixed-random editing improves mean ranking at every post-initial round, reaching
$\Delta z=0.056$ at $672$ labels (Figure~\ref{fig:main}a,b). Fixed-UCB editing instead reaches
$\Delta R_{96}=0.064$ versus no edits. These trajectories combine editing with additional
measurements; they do not isolate reflection.

\paragraph{Repeated feedback has no detected advantage over one recommendation.}
Full LLM minus one-call control is $-0.011$ in $z$ ($[-0.028,+0.007]$) and $+0.0005$ in
$R_{96}$ ($[-0.012,+0.013]$). Replacing the diagnostic report with a scalar score likewise
changes mean ranking by only $+0.0006$. These results do not establish equivalence, but they
provide no evidence that repeated rich feedback explains the full controller's performance.
They also leave open whether feedback helps the successful fixed-acquisition arms, for which
a matched one-call ablation was not run.

\paragraph{The feedback signal depends on which labels are acquired.}
A post-hoc configuration screen at $672$ random labels finds a CV-selected gain of $0.072$
in test $z$ over the default ($32/36$ assays), showing that the main null result does not
reflect an already-optimal starting recipe. Yet the CV--test relationship varies with
acquisition: the mean CV-minus-test gap is $-0.032$ for no edits and $-0.184$ for fixed
diversity-greedy acquisition. These negative gaps indicate lower CV than test scores, not
optimistic CV bias. They motivate investigating distribution-aware feedback, but do not
establish that distribution shift causes the full controller's ranking loss. The screen and
diagnostics are descriptive rather than matched causal tests (Appendix~\ref{app:diagnostics}).

\paragraph{Controller and split sensitivity qualify the findings.}
With the legal-value menu omitted, Opus~5 exceeds Sonnet~5 by $0.035$ in $z$
($[0.026,0.044]$, $q=7.1\times10^{-8}$; $35/36$ wins) and Haiku~4.5 by $0.030$
($[0.019,0.040]$, $q<10^{-3}$). These are deployed-controller comparisons: schema rejections
and failed calls also affect behavior (Appendix~\ref{app:diagnostics}). On the position-disjoint
split, full LLM versus random edits remains inconclusive ($\Delta z=0.023$,
$[-0.003,+0.048]$); random-fold CV does not match this test distribution.
Five-seed sensitivity retains an inconclusive primary ranking contrast and a positive
fixed-UCB editing effect on recovery (Appendix~\ref{app:sensitivity}).

\section{Conclusion and limitation}
\label{sec:conclusion}
At $96$ labels per round, LLM-guided specialist editing can help under a prescribed acquisition
rule, but the observed benefit depends on whether the goal is global ranking or top-variant
recovery. The full controller shows no detected ranking advantage over random edits, and
repeated feedback shows no detected incremental value over one recommendation. These findings
support evaluating model proposals, acquisition, and reflection as separate contributions.
Metric-dependent trade-offs also arise in credit-card fraud detection, where stronger
amount weighting can improve dollar recall while reducing ranking quality and dollar
precision \citep{wu2026classweightingversusconditioning}.

The positive ablations are exploratory. Our evidence is limited to offline single-substitution
assays, one frozen encoder, a bounded edit schema, and six acquisition rounds; it does not
establish gains in prospective wet-lab discovery. Only one label budget is tested, so a causal
budget threshold cannot be inferred. Repeated selection on purchased-label CV and changing
acquisition distributions may affect the gate, but the present comparisons do not isolate that
mechanism. Matched one-call controls for the fixed-acquisition arms and evaluation under
distribution-aware selection are the next tests needed to establish the value of reflection.

\clearpage
\bibliographystyle{plainnat}
\bibliography{refs}

@article{lin2023esm2,
author = {Zeming Lin  and Halil Akin  and Roshan Rao  and Brian Hie  and Zhongkai Zhu  and Wenting Lu  and Nikita Smetanin  and Robert Verkuil  and Ori Kabeli  and Yaniv Shmueli  and Allan dos Santos Costa  and Maryam Fazel-Zarandi  and Tom Sercu  and Salvatore Candido  and Alexander Rives },
title = {Evolutionary-scale prediction of atomic-level protein structure with a language model},
journal = {Science},
volume = {379},
number = {6637},
pages = {1123-1130},
year = {2023},
doi = {10.1126/science.ade2574},
URL = {https://www.science.org/doi/abs/10.1126/science.ade2574},
eprint = {https://www.science.org/doi/pdf/10.1126/science.ade2574}}

@inproceedings{meier2021esm1v,
 author = {Meier, Joshua and Rao, Roshan and Verkuil, Robert and Liu, Jason and Sercu, Tom and Rives, Alex},
 booktitle = {Advances in Neural Information Processing Systems},
 editor = {M. Ranzato and A. Beygelzimer and Y. Dauphin and P.S. Liang and J. Wortman Vaughan},
 pages = {29287--29303},
 publisher = {Curran Associates, Inc.},
 title = {Language models enable zero-shot prediction of the effects of mutations on protein function},
 url = {https://proceedings.neurips.cc/paper_files/paper/2021/file/f51338d736f95dd42427296047067694-Paper.pdf},
 volume = {34},
 year = {2021}
}

@inproceedings{notin2023proteingym,
 author = {Notin, Pascal and Kollasch, Aaron and Ritter, Daniel and van Niekerk, Lood and Paul, Steffanie and Spinner, Han and Rollins, Nathan and Shaw, Ada and Orenbuch, Rose and Weitzman, Ruben and Frazer, Jonathan and Dias, Mafalda and Franceschi, Dinko and Gal, Yarin and Marks, Debora},
 booktitle = {Advances in Neural Information Processing Systems},
 doi = {10.52202/075280-2810},
 editor = {A. Oh and T. Naumann and A. Globerson and K. Saenko and M. Hardt and S. Levine},
 pages = {64331--64379},
 publisher = {Curran Associates, Inc.},
 title = {ProteinGym: Large-Scale Benchmarks for Protein Fitness Prediction and Design},
 url = {https://proceedings.neurips.cc/paper_files/paper/2023/file/cac723e5ff29f65e3fcbb0739ae91bee-Paper-Datasets_and_Benchmarks.pdf},
 volume = {36},
 year = {2023}
}

@article{hsu2022learning,
  title   = {Learning protein fitness models from evolutionary and assay-labeled data},
  author  = {Hsu, Chloe and Nisonoff, Hunter and Fannjiang, Clara and Listgarten, Jennifer},
  journal = {Nature Biotechnology},
  volume  = {40},
  number  = {7},
  pages   = {1114--1122},
  year    = {2022},
  doi     = {10.1038/s41587-021-01146-5}
}

@article{jiang2025evolvepro,
author = {Kaiyi Jiang  and Zhaoqing Yan  and Matteo Di Bernardo  and Samantha R. Sgrizzi  and Lukas Villiger  and Alisan Kayabolen  and B. J. Kim  and Josephine K. Carscadden  and Masahiro Hiraizumi  and Hiroshi Nishimasu  and Jonathan S. Gootenberg  and Omar O. Abudayyeh },
title = {Rapid in silico directed evolution by a protein language model with EVOLVEpro},
journal = {Science},
volume = {387},
number = {6732},
pages = {eadr6006},
year = {2025},
doi = {10.1126/science.adr6006},
URL = {https://www.science.org/doi/abs/10.1126/science.adr6006},
eprint = {https://www.science.org/doi/pdf/10.1126/science.adr6006}}

@inproceedings{roohani2025biodiscoveryagent,
title={BioDiscoveryAgent: An {AI} Agent for Designing Genetic Perturbation Experiments},
author={Yusuf H Roohani and Andrew H. Lee and Qian Huang and Jian Vora and Zachary Steinhart and Kexin Huang and Alexander Marson and Percy Liang and Jure Leskovec},
booktitle={The Thirteenth International Conference on Learning Representations},
year={2025},
url={https://openreview.net/forum?id=HAwZGLcye3}
}

@inproceedings{agrawal2026gepa,
title={{GEPA}: Reflective Prompt Evolution Can Outperform Reinforcement Learning},
author={Lakshya A Agrawal and Shangyin Tan and Dilara Soylu and Noah Ziems and Rishi Khare and Krista Opsahl-Ong and Arnav Singhvi and Herumb Shandilya and Michael J Ryan and Meng Jiang and Christopher Potts and Koushik Sen and Alex Dimakis and Ion Stoica and Dan Klein and Matei Zaharia and Omar Khattab},
booktitle={The Fourteenth International Conference on Learning Representations},
year={2026},
url={https://openreview.net/forum?id=RQm2KQTM5r}
}

@inproceedings{yang2024large,
title={Large Language Models as Optimizers},
author={Chengrun Yang and Xuezhi Wang and Yifeng Lu and Hanxiao Liu and Quoc V Le and Denny Zhou and Xinyun Chen},
booktitle={The Twelfth International Conference on Learning Representations},
year={2024},
url={https://openreview.net/forum?id=Bb4VGOWELI}
}

@inproceedings{zhang2026darwin,
title={Darwin G\"odel Machine: Open-Ended Evolution of Self-Improving Agents},
author={Jenny Zhang and Shengran Hu and Cong Lu and Robert Tjarko Lange and Jeff Clune},
booktitle={The Fourteenth International Conference on Learning Representations},
year={2026},
url={https://openreview.net/forum?id=pUpzQZTvGY}
}

@article{huang2026agent,
  title = {{CE-Agent}: a cloud-edge collaborative agent framework with dynamic offloading and synergistic distillation},
  author = {Huang, Yan and Zhang, Meimei and Lei, Hong and Ji, Chenyang and Wang, Jiayi and Zhu, Huidong},
  journal = {Journal of Cloud Computing},
  year = {2026},
  publisher = {Springer},
  doi = {10.1186/s13677-026-00984-5},
  url = {https://doi.org/10.1186/s13677-026-00984-5}
}

@inproceedings{liang2025revealing,
  title = {Treatment Non-Adherence Bias in Clinical Machine Learning: A Real-World Study on Hypertension Medication},
  author = {Liang, Zhongyuan and Suresh, Arvind and Chen, Irene Y.},
  booktitle = {Proceedings of the sixth Conference on Health, Inference, and Learning},
  series = {Proceedings of Machine Learning Research},
  volume = {287},
  pages = {430--442},
  year = {2025},
  publisher = {PMLR},
  url = {https://proceedings.mlr.press/v287/liang25a.html}
}

@phdthesis{li2022root,
  title = {Root Cause Analysis of Failure in Molecular Biology Workflows},
  author = {Li, Shuowei},
  school = {University of Washington},
  year = {2022},
  url = {https://hdl.handle.net/1773/49344}
}

@misc{wang2026embeddingfoundationmodelpredictions,
  title = {Embedding Foundation Model Predictions in Discrete-Choice Models with Structural Guarantees},
  author = {Wang, Yingshuo and Sun, Xian and Li, Yanhang and Fan, Zhichao and Zhuang, Zexin},
  year = {2026},
  eprint = {2606.26432},
  archivePrefix = {arXiv},
  primaryClass = {cs.LG},
  url = {https://arxiv.org/abs/2606.26432}
}

@inproceedings{yuan2022opticalflow,
  title = {Optical Flow Training under Limited Label Budget via Active Learning},
  author = {Yuan, Shuai and Sun, Xian and Kim, Hannah and Yu, Shuzhi and Tomasi, Carlo},
  booktitle = {Computer Vision -- ECCV 2022},
  pages = {410--427},
  year = {2022},
  publisher = {Springer},
  doi = {10.1007/978-3-031-20047-2_24},
  url = {https://www.ecva.net/papers/eccv_2022/papers_ECCV/html/4388_ECCV_2022_paper.php}
}

@inproceedings{sun2025audioenhanced,
  title = {Audio-Enhanced Vision-Language Modeling with Latent Space Broadening for High Quality Data Expansion},
  author = {Sun, Yu and Li, Yin and Sun, Ruixiao and Liu, Chunhui and Zhou, Fangming and Jin, Ze and Wang, Linjie and Shen, Xiang and Hao, Zhuolin and Xiong, Hongyu},
  booktitle = {Proceedings of the 31st ACM SIGKDD Conference on Knowledge Discovery and Data Mining V.2},
  year = {2025},
  pages = {4872--4881},
  publisher = {Association for Computing Machinery},
  doi = {10.1145/3711896.3737195},
  url = {https://doi.org/10.1145/3711896.3737195}
}

@article{wu2026classweightingversusconditioning,
  title = {Class Weighting versus Amount Conditioning in Credit-Card Fraud Detection: A Dollar-Metric Study with a Temporal Explanation Audit},
  author = {Wu, Chenyu},
  journal = {arXiv preprint arXiv:2607.14686},
  year = {2026},
  eprint = {2607.14686},
  archivePrefix = {arXiv},
  primaryClass = {cs.CE},
  doi = {10.48550/arXiv.2607.14686},
  url = {https://arxiv.org/abs/2607.14686}
}
\clearpage
\appendix
\section{Protocol and reproducibility}
\label{app:protocol}
\paragraph{Available evidence and reconstruction.}
All reported statistics come from a frozen results snapshot containing per-round and per-run
endpoints, supporting diagnostic summaries, and a provenance manifest.
The snapshot cutoff is 27 September 2026, 05:04:52 UTC. An analysis script reconstructs the
paired comparisons, figures, and tables from the exported endpoint summaries without modifying
the snapshot, and a separate validation script checks input hashes, label budgets, round
completeness, common initial endpoints, endpoint aggregation, and agreement with the supplied
comparison statistics.
Raw assay labels, embeddings, complete run records, and controller-call logs are not part of
this summary snapshot; the original server audit cannot be rerun from these summaries alone.

An arm is eligible only if every one of the $36$ assays has all requested seeds and all seven
recorded states (initialization and rounds 1--6). Eligibility does not depend on effect size,
significance, or direction. The snapshot contains $4{,}211$ unique completed run summaries and
$29{,}477$ round records. The three-seed analyses include $20$ random-split arms and $11$
position-split arms. The five-seed analysis includes $11$ complete random-split arms; nominal
$5\times$ search is excluded there because one of its $180$ required runs is absent.
All eligible arms are reported below, including controls omitted from the main table.

The original analysis code specifies the primary contrast and equivalence margin.
We retain this three-seed test as the primary inference; subsequent ablations and expanded
seed analyses remain exploratory. BH correction uses the original declared comparison families:
$84$ contrasts per endpoint for the three-seed random split, $18$ for the position split,
and $21$ for the five-seed random split. The primary contrast retains its unadjusted inference;
all reported $q$ values use the complete family, including comparisons absent from the main text.
Intervals are pointwise $95\%$ paired $t$ intervals, not simultaneous confidence bands.

\paragraph{Arm definitions.}
The principal LLM arms share a preamble describing the loop and gate. Full feedback includes
CV Spearman with a bootstrap interval, errors by residue position, amino-acid class and fitness
quintile, uncertainty calibration, configuration, edit history, and legal action values.
Scalar feedback retains the menu but replaces the diagnostic report with CV Spearman.
No-menu feedback retains diagnostics and current field names but omits enumerated legal values.

One-call control makes its only proposal in round~1. After validation and the ordinary CV gate,
its resulting configuration remains fixed while the predictor is retrained as labels accumulate.
No-curation control permits pseudo-labeling, feature, training, and acquisition edits only.
Fixed-random and fixed-UCB model-editing arms permit all non-acquisition edits while prescribing
the acquisition rule each round. Acquisition-only LLM control changes the acquisition strategy
and exploration parameter while retaining the other defaults. Non-LLM acquisition controls
prescribe greedy, UCB, or diversity-greedy selection and leave the remaining fields at default.
Static ridge uses an augmented recipe inspired by \citet{hsu2022learning} and re-tunes regularization.

The configured model names are Opus~5, Sonnet~5, and Haiku~4.5. Their CLI identifiers are
\path{claude-opus-5}, \path{claude-sonnet-5}, and
\path{claude-haiku-4-5-20251001}, respectively.
These names describe the experiment configurations, not independently verified model sizes.
The prompt templates and selected response examples are reproduced verbatim in
Appendix~\ref{app:prompts}.
Controller substitutions use the same feedback variants, label budgets, and edit schema.

\paragraph{Endpoint definition.}
The exported \texttt{topk} field is computed by the evaluation code as follows:
\[
 R_{96}=\frac{|P_{96}\cap T_m|}{\min(96,m)},\qquad
 m=\max\!\left(1,\operatorname{round}(0.1n_{\mathrm{test}})\right).
\]
This denominator is the maximum possible overlap, so $R_{96}$ is bounded by one and equals
recall only for $m\leq96$. Both endpoints use sealed-test predictions; $R_{96}$ is not the
acquired batch's measured hit rate. We average rounds 3--6 ($384$--$672$ labels), then seeds
within assays, giving each assay equal weight.

\begin{figure}[tb]
\centering
\includegraphics[width=\textwidth]{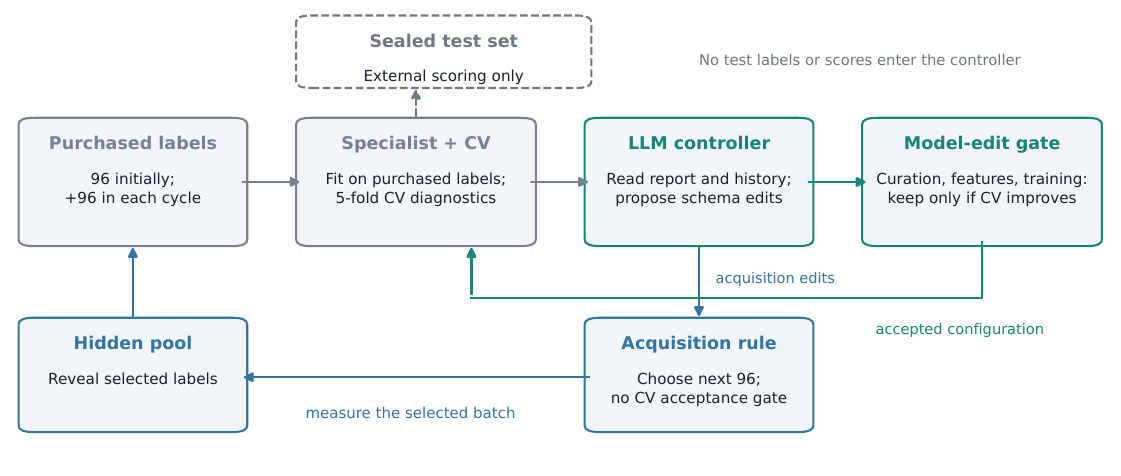}
\caption{Information flow in the budgeted loop. Purchased-label CV informs model edits and
the controller. Non-acquisition edits pass an acceptance gate; acquisition follows a separate
path and reveals the next $96$ labels. The sealed test set is used only for external scoring.}
\label{fig:system}
\end{figure}

\section{Diagnostic evidence and interpretation}
\label{app:diagnostics}
\paragraph{Available headroom.}
A post-hoc configuration screen holds a random set of $672$ labels fixed, samples $120$
configurations per assay, and selects by three-fold CV. Candidate heads are ridge, a small MLP,
and a ridge ensemble; gradient boosting is excluded for computational cost. Fourteen configurations
with nonfinite correlations from constant predictions on CCDB are omitted from the summaries.
The CV-selected configuration gains a mean $0.072$ test Fisher $z$ over the default and improves
$32/36$ assays. Selection uses CV, not test performance. This shows that the default is not
already optimal, but different compute, CV folds, and labeled-data trajectories prevent treating
the screen as a matched online baseline or a formal upper bound.

\paragraph{Acquisition and the feedback distribution.}
In the supplied diagnostics, the mean CV-minus-test Fisher-$z$ gap over rounds 3--6 is $-0.032$
for no edits, $-0.068$ for fixed UCB, $-0.184$ for diversity-greedy, and $-0.152$ for full LLM
control. A negative gap means CV is lower than sealed-test performance; it is not evidence of
optimistically biased CV. The variation is consistent with acquisition-dependent evaluation
distributions, but does not identify why full control loses ranking performance.
Changing acquisition also changes subsequent training sets and proposals. Testing a
distribution-aware gate would require a new controlled comparison.

\paragraph{Schema behavior and controller reliability.}
The Opus no-menu arm reaches mean $z=0.638$, versus $0.603$ for no edits. However, its validator
rejects $640$ categorical entries across $509/648$ calls, and the arm never changes the default
head or feature set. Its apparent benefit therefore cannot isolate the value of removing the
menu: the intervention also changes the effective set of edits.

All four Sonnet and Haiku substitution arms cover $36$ assays and three seeds. Under full
feedback, Opus exceeds each substitute by approximately $0.022$ in $z$ ($q=0.019$ for Sonnet,
$q=0.008$ for Haiku). Under no-menu feedback, mean $z$ is $0.638$ for Opus, $0.602$ for Sonnet,
and $0.608$ for Haiku. Across the two Haiku arms, $190/1{,}296$ scheduled calls fail after
wrapper retries; the two Sonnet arms have no failed calls. Among the Opus ablations, there are
$26/648$ failed calls under fixed random acquisition, $33/648$ without curation, and $5/108$
for one-call control. Failed calls preserve the current configuration.
These comparisons measure deployed systems, combining model behavior, schema compatibility,
and execution reliability.

\begin{samepage}
\section{All completed three-seed controls}
\label{app:controls}
The tables report the mean Fisher-transformed rank correlation, its difference from no edits,
and normalized recovery. Each mean gives equal weight to the $36$ assays after averaging
seeds $0$--$2$ and rounds 3--6. On the position split, random-fold CV operates on purchased
variants while the test set consists of unseen residue positions; the two evaluation
distributions consequently differ.
\end{samepage}

\begin{table}[htbp]
\centering
\caption{Random split: all $20$ complete arms, $108$ runs per arm.
Unqualified LLM arms use the configured Opus~5 controller.}
\label{tab:all_random}
\small
\begin{tabular}{@{}lrrr@{}}
\toprule
Controller & Mean $z$ & $\Delta z$ vs. no edits & $R_{96}$ \\
\midrule
Fixed div.-greedy & 0.597 & $-0.007$ & 0.384 \\
Fixed greedy & 0.566 & $-0.038$ & 0.380 \\
Fixed UCB & 0.589 & $-0.014$ & 0.383 \\
No edits & 0.603 & $+0.000$ & 0.349 \\
LLM acq. only & 0.591 & $-0.012$ & 0.377 \\
LLM full & 0.590 & $-0.014$ & 0.391 \\
Full: Haiku 4.5 & 0.568 & $-0.036$ & 0.375 \\
Full: Sonnet 5 & 0.568 & $-0.036$ & 0.384 \\
LLM + fixed UCB & 0.576 & $-0.028$ & 0.406 \\
LLM + random acq. & 0.647 & $+0.043$ & 0.358 \\
LLM no curation & 0.598 & $-0.006$ & 0.390 \\
LLM no menu & 0.638 & $+0.034$ & 0.375 \\
No menu: Haiku 4.5 & 0.608 & $+0.005$ & 0.350 \\
No menu: Sonnet 5 & 0.602 & $-0.001$ & 0.364 \\
LLM one call & 0.600 & $-0.003$ & 0.390 \\
LLM scalar & 0.590 & $-0.013$ & 0.403 \\
Random edits & 0.597 & $-0.006$ & 0.359 \\
Search 5x & 0.623 & $+0.020$ & 0.354 \\
Search 1x & 0.619 & $+0.016$ & 0.352 \\
Static ridge & 0.599 & $-0.004$ & 0.333 \\
\bottomrule
\end{tabular}

\end{table}
\clearpage
\begin{table}[htbp]
\centering
\caption{Position-disjoint split: all $11$ complete arms, $108$ runs per arm.}
\label{tab:all_position}
\small
\begin{tabular}{@{}lrrr@{}}
\toprule
Controller & Mean $z$ & $\Delta z$ vs. no edits & $R_{96}$ \\
\midrule
Fixed div.-greedy & 0.402 & $+0.000$ & 0.285 \\
Fixed greedy & 0.404 & $+0.003$ & 0.288 \\
Fixed UCB & 0.402 & $+0.001$ & 0.285 \\
No edits & 0.401 & $+0.000$ & 0.276 \\
LLM full & 0.461 & $+0.060$ & 0.319 \\
LLM no menu & 0.453 & $+0.052$ & 0.293 \\
LLM scalar & 0.446 & $+0.045$ & 0.316 \\
Random edits & 0.439 & $+0.037$ & 0.290 \\
Search 5x & 0.506 & $+0.105$ & 0.313 \\
Search 1x & 0.499 & $+0.098$ & 0.302 \\
Static ridge & 0.492 & $+0.091$ & 0.296 \\
\bottomrule
\end{tabular}

\end{table}

\section{Five-seed sensitivity}
\label{app:sensitivity}
Expanding to seeds $0$--$4$ leaves full LLM versus random edits inconclusive for ranking:
$\Delta z=-0.012$ ($95\%$ CI $[-0.034,+0.010]$). The fixed-UCB model-editing arm retains a
recovery gain over UCB alone, $\Delta R_{96}=0.020$ ($[0.007,0.033]$, $q=0.005$).
This is a post-hoc sensitivity analysis, not a replacement for the three-seed primary test.
The fixed-random editing and one-call arms lack complete five-seed grids, so this analysis
does not establish their robustness to additional seeds.
\begin{table}[htbp]
\centering
\caption{All $11$ complete random-split arms with seeds $0$--$4$, $180$ runs per arm,
averaged over rounds 3--6. Eligibility follows Appendix~\ref{app:protocol}.}
\label{tab:five_seeds}
\small
\begin{tabular}{@{}lrrr@{}}
\toprule
Controller & Mean $z$ & $\Delta z$ vs. no edits & $R_{96}$ \\
\midrule
Fixed div.-greedy & 0.591 & $-0.011$ & 0.383 \\
Fixed greedy & 0.564 & $-0.038$ & 0.382 \\
Fixed UCB & 0.584 & $-0.018$ & 0.385 \\
No edits & 0.602 & $+0.000$ & 0.354 \\
LLM full & 0.587 & $-0.015$ & 0.392 \\
LLM + fixed UCB & 0.573 & $-0.029$ & 0.405 \\
LLM no menu & 0.635 & $+0.032$ & 0.375 \\
LLM scalar & 0.584 & $-0.019$ & 0.402 \\
Random edits & 0.599 & $-0.003$ & 0.360 \\
Search 1x & 0.614 & $+0.012$ & 0.352 \\
Static ridge & 0.596 & $-0.006$ & 0.334 \\
\bottomrule
\end{tabular}

\end{table}

\clearpage
\section{Action space and assay list}
\label{app:actions}
The schema rejects unknown fields and illegal categories and clamps out-of-range numbers.
Acquisition comprises the selection rule and its exploration parameter; the remaining groups
pass through the shared CV gate. The schema and defaults below reproduce the original settings.
\begingroup
\footnotesize
\setlength{\tabcolsep}{3pt}
\renewcommand{\arraystretch}{1.12}
\begin{longtable}{@{}>{\raggedright\arraybackslash}p{0.32\textwidth}>{\raggedright\arraybackslash}p{0.51\textwidth}>{\raggedright\arraybackslash}p{0.13\textwidth}@{}}
\caption{Allowed edits and default values. Numeric ranges are inclusive.}\label{tab:actions}\\
\toprule
Field & Allowed values & Default \\
\midrule
\endfirsthead
\multicolumn{3}{l}{\textit{Action space (continued)}}\\
\toprule
Field & Allowed values & Default \\
\midrule
\endhead
\bottomrule
\endfoot
\multicolumn{3}{@{}l}{\textbf{Curation}} \\
drop\_\allowbreak{}outliers.method & none, iqr, zscore & none \\
drop\_\allowbreak{}outliers.k & [1.0, 4.0] (real) & 3.0 \\
dedup\_\allowbreak{}by\_\allowbreak{}position & False, True & False \\
reweight & none, inverse\_\allowbreak{}zs\_\allowbreak{}confidence, fitness\_\allowbreak{}decile\_\allowbreak{}balanced & none \\
clip\_\allowbreak{}targets & none, winsorize\_\allowbreak{}1, winsorize\_\allowbreak{}5 & none \\
\multicolumn{3}{@{}l}{\textbf{Pseudo-labeling and features}} \\
pseudo\_\allowbreak{}label & none, zero\_\allowbreak{}shot\_\allowbreak{}topk, self\_\allowbreak{}training\_\allowbreak{}confident & none \\
n & [0, 2000] (integer) & 0 \\
feature\_\allowbreak{}set & emb, emb+zs, emb+zs+onehot, zs+onehot & emb+zs \\
\multicolumn{3}{@{}l}{\textbf{Acquisition}} \\
strategy & random, greedy, ucb, ensemble\_\allowbreak{}disagreement, diversity\_\allowbreak{}greedy, zero\_\allowbreak{}shot\_\allowbreak{}prior & random \\
beta & [0.0, 3.0] (real) & 1.0 \\
\multicolumn{3}{@{}l}{\textbf{Training recipe}} \\
head & ridge, mlp\_\allowbreak{}small, ensemble\_\allowbreak{}ridge, gbr & ridge \\
loss & mse, huber, rank & mse \\
alpha & [0.001, 1000.0] (real) & 1.0 \\
epochs & [20, 200] (integer) & 200 \\
standardize & False, True & True \\
\end{longtable}
\endgroup

\clearpage
\subsection{Assays}
\label{app:assays}
The pre-main-result filter selects the following $36$ assays, totaling $101{,}201$
single-substitution variants. Virus names are abbreviated for readability; the full
taxonomic strings are given in the ProteinGym reference table. \emph{Fitness} denotes the
OrganismalFitness readout category.
\begingroup
\fontsize{8.5}{9.5}\selectfont
\setlength{\tabcolsep}{3pt}
\renewcommand{\arraystretch}{1.04}
\begin{longtable}{@{}>{\raggedright\arraybackslash}p{0.42\textwidth}rr>{\raggedright\arraybackslash}p{0.12\textwidth}>{\raggedright\arraybackslash}p{0.22\textwidth}@{}}
\caption{Assay metadata. Length counts residues; variants are single substitutions.}\label{tab:assays}\\
\toprule
Assay & Length & Variants & Readout & Organism \\
\midrule
\endfirsthead
\multicolumn{5}{l}{\textit{Assay metadata (continued)}}\\
\toprule
Assay & Length & Variants & Readout & Organism \\
\midrule
\endhead
\bottomrule
\endfoot
A0A247D711\_\allowbreak{}LISMN\_\allowbreak{}Stadelmann\_\allowbreak{}2021 & 87 & 1,653 & Activity & Listeria monocytogenes \\
A4GRB6\_\allowbreak{}PSEAI\_\allowbreak{}Chen\_\allowbreak{}2020 & 266 & 5,004 & Fitness & Pseudomonas aeruginosa \\
AACC1\_\allowbreak{}PSEAI\_\allowbreak{}Dandage\_\allowbreak{}2018 & 177 & 1,801 & Fitness & Pseudomonas aeruginosa \\
BLAT\_\allowbreak{}ECOLX\_\allowbreak{}Firnberg\_\allowbreak{}2014 & 286 & 4,783 & Fitness & Escherichia coli \\
CALM1\_\allowbreak{}HUMAN\_\allowbreak{}Weile\_\allowbreak{}2017 & 149 & 1,813 & Fitness & Homo sapiens \\
CASP3\_\allowbreak{}HUMAN\_\allowbreak{}Roychowdhury\_\allowbreak{}2020 & 258 & 1,567 & Activity & Homo sapiens \\
CASP7\_\allowbreak{}HUMAN\_\allowbreak{}Roychowdhury\_\allowbreak{}2020 & 281 & 1,680 & Activity & Homo sapiens \\
CCDB\_\allowbreak{}ECOLI\_\allowbreak{}Tripathi\_\allowbreak{}2016 & 101 & 1,663 & Fitness & Escherichia coli \\
DYR\_\allowbreak{}ECOLI\_\allowbreak{}Thompson\_\allowbreak{}2019 & 159 & 2,363 & Fitness & Escherichia coli \\
ESTA\_\allowbreak{}BACSU\_\allowbreak{}Nutschel\_\allowbreak{}2020 & 212 & 2,172 & Stability & Bacillus subtilis \\
HEM3\_\allowbreak{}HUMAN\_\allowbreak{}Loggerenberg\_\allowbreak{}2023 & 361 & 5,689 & Activity & Homo sapiens \\
KCNE1\_\allowbreak{}HUMAN\_\allowbreak{}Muhammad\_\allowbreak{}2023\_\allowbreak{}function & 129 & 2,315 & Activity & Homo sapiens \\
KKA2\_\allowbreak{}KLEPN\_\allowbreak{}Melnikov\_\allowbreak{}2014 & 264 & 4,960 & Fitness & Klebsiella pneumoniae \\
MET\_\allowbreak{}HUMAN\_\allowbreak{}Estevam\_\allowbreak{}2023 & 287 & 5,393 & Activity & Homo sapiens \\
MLAC\_\allowbreak{}ECOLI\_\allowbreak{}MacRae\_\allowbreak{}2023 & 211 & 4,007 & Fitness & Escherichia coli \\
MTH3\_\allowbreak{}HAEAE\_\allowbreak{}RockahShmuel\_\allowbreak{}2015 & 330 & 1,777 & Fitness & Haemophilus aegyptius \\
NUD15\_\allowbreak{}HUMAN\_\allowbreak{}Suiter\_\allowbreak{}2020 & 164 & 2,844 & Expression & Homo sapiens \\
OTC\_\allowbreak{}HUMAN\_\allowbreak{}Lo\_\allowbreak{}2023 & 354 & 1,570 & Activity & Homo sapiens \\
P84126\_\allowbreak{}THETH\_\allowbreak{}Chan\_\allowbreak{}2017 & 254 & 1,519 & Fitness & Thermus thermophilus \\
PHOT\_\allowbreak{}CHLRE\_\allowbreak{}Chen\_\allowbreak{}2023 & 118 & 2,122 & Activity & Chlamydomonas reinhardtii \\
Q53Z42\_\allowbreak{}HUMAN\_\allowbreak{}McShan\_\allowbreak{}2019\_\allowbreak{}binding-TAPBPR & 365 & 3,344 & Binding & Homo sapiens \\
R1AB\_\allowbreak{}SARS2\_\allowbreak{}Flynn\_\allowbreak{}2022 & 306 & 5,725 & Fitness & SARS-CoV-2 \\
RASH\_\allowbreak{}HUMAN\_\allowbreak{}Bandaru\_\allowbreak{}2017 & 189 & 3,134 & Activity & Homo sapiens \\
RASK\_\allowbreak{}HUMAN\_\allowbreak{}Weng\_\allowbreak{}2022\_\allowbreak{}abundance & 188 & 3,066 & Expression & Homo sapiens \\
REV\_\allowbreak{}HV1H2\_\allowbreak{}Fernandes\_\allowbreak{}2016 & 116 & 2,147 & Fitness & HIV-1 (HXB2) \\
RNC\_\allowbreak{}ECOLI\_\allowbreak{}Weeks\_\allowbreak{}2023 & 226 & 4,277 & Activity & Escherichia coli \\
SERC\_\allowbreak{}HUMAN\_\allowbreak{}Xie\_\allowbreak{}2023 & 370 & 1,914 & Fitness & Homo sapiens \\
SUMO1\_\allowbreak{}HUMAN\_\allowbreak{}Weile\_\allowbreak{}2017 & 101 & 1,700 & Fitness & Homo sapiens \\
SYUA\_\allowbreak{}HUMAN\_\allowbreak{}Newberry\_\allowbreak{}2020 & 140 & 2,497 & Fitness & Homo sapiens \\
TAT\_\allowbreak{}HV1BR\_\allowbreak{}Fernandes\_\allowbreak{}2016 & 86 & 1,577 & Fitness & HIV-1 (BRU/LAI) \\
TPK1\_\allowbreak{}HUMAN\_\allowbreak{}Weile\_\allowbreak{}2017 & 243 & 3,181 & Fitness & Homo sapiens \\
TPMT\_\allowbreak{}HUMAN\_\allowbreak{}Matreyek\_\allowbreak{}2018 & 245 & 3,648 & Expression & Homo sapiens \\
TRPC\_\allowbreak{}SACS2\_\allowbreak{}Chan\_\allowbreak{}2017 & 248 & 1,519 & Fitness & Thermus thermophilus \\
TRPC\_\allowbreak{}THEMA\_\allowbreak{}Chan\_\allowbreak{}2017 & 252 & 1,519 & Fitness & Thermus thermophilus \\
UBC9\_\allowbreak{}HUMAN\_\allowbreak{}Weile\_\allowbreak{}2017 & 159 & 2,563 & Fitness & Homo sapiens \\
VKOR1\_\allowbreak{}HUMAN\_\allowbreak{}Chiasson\_\allowbreak{}2020\_\allowbreak{}abundance & 163 & 2,695 & Expression & Homo sapiens \\
\end{longtable}
\endgroup

\clearpage
\section{Controller prompts and example responses}
\label{app:prompts}
\subsection{Prompt templates}
\label{app:prompt-templates}
Verbatim templates. Each call concatenates the shared preamble, the diagnostic report (or, under
scalar feedback, CV Spearman alone), the current configuration, the action menu (full and scalar
feedback only) and the instruction for its variant.

\paragraph{Shared preamble (all principal LLM arms).}
\begin{lstlisting}
You are improving a small supervised model that predicts a measured fitness value for
single amino-acid substitutions in one protein, from a public deep mutational scanning dataset.

The loop works like this. A fixed set of variants has hidden measurements. Each round you may change
the data curation, the pseudo-label generation, the training recipe, and the rule that picks which 96
still-unmeasured variants are measured next. The model is then retrained on all measured labels and
re-evaluated by 5-fold cross-validation over the labels purchased so far. Your edit is kept only if
that cross-validated Spearman improves; otherwise it is reverted. Changes to the acquisition rule are
always applied, because a batch of measurements cannot be un-ordered.

You never see the held-out test set and you never see test metrics. Optimizing the validation number
by exploiting its noise will not help: the reported outcome is measured on data you cannot see.
\end{lstlisting}

\paragraph{Action menu (full and scalar feedback; withheld under no-menu feedback).}
\begin{lstlisting}
You may change any subset of the following configuration fields. These are the ONLY legal
fields and the ONLY legal values. Anything else is rejected.

curation.drop_outliers.method : one of "none", "iqr", "zscore"
curation.drop_outliers.k      : float in [1.0, 4.0]      (threshold for the chosen method)
curation.dedup_by_position    : true or false            (keep at most one labeled variant per residue position)
curation.reweight             : one of "none", "inverse_zs_confidence", "fitness_decile_balanced"
curation.clip_targets         : one of "none", "winsorize_1", "winsorize_5"
generation.pseudo_label       : one of "none", "zero_shot_topk", "self_training_confident"
generation.n                  : int in [0, 2000]         (number of pseudo-labeled variants to add)
generation.feature_set        : one of "emb", "emb+zs", "emb+zs+onehot", "zs+onehot"
acquisition.strategy          : one of "random", "greedy", "ucb", "ensemble_disagreement",
                                "diversity_greedy", "zero_shot_prior"
acquisition.beta              : float in [0.0, 3.0]      (exploration weight for ucb)
recipe.head                   : one of "ridge", "mlp_small", "ensemble_ridge", "gbr"
recipe.loss                   : one of "mse", "huber", "rank"   ("huber" applies to the linear head;
                                "rank" fits on rank-transformed targets and applies to any head)
recipe.alpha                  : float in [0.001, 1000]   (regularization strength)
recipe.epochs                 : int in [20, 200]         (affects only the iterative heads,
                                mlp_small and gbr)
recipe.standardize            : true or false

Field meanings: "emb" is a mean-pooled frozen protein language model embedding of the full mutant
sequence; "zs" is that model's zero-shot log-likelihood ratio for the substitution; "onehot" is a
one-hot encoding of the substitution. Acquisition chooses which 96 unlabeled variants get measured
next round.
\end{lstlisting}

\paragraph{Instruction, full and scalar feedback.}
\begin{lstlisting}
Decide what to change for the next round. Think about what the report actually shows: whether the
model is overfitting, whether any stratum is systematically worse, whether the uncertainty estimate
is trustworthy enough to drive an uncertainty-based acquisition rule, and whether the remaining label
budget is better spent exploring or exploiting.

Reply with a short justification (at most 5 sentences) and then a single JSON object containing ONLY
the fields you want to change. To change nothing, return {}.
\end{lstlisting}

\paragraph{Instruction, no-menu feedback.}
\begin{lstlisting}
Decide what to change for the next round. Think about what the report actually shows: whether the
model is overfitting, whether any stratum is systematically worse, whether the uncertainty estimate
is trustworthy enough to drive an uncertainty-based acquisition rule, and whether the remaining label
budget is better spent exploring or exploiting.

Reply with a short justification (at most 5 sentences) and then a single JSON object containing ONLY
the configuration fields you want to change, using the same structure as the configuration shown
above. To change nothing, return {}.
\end{lstlisting}

\subsection{Representative controller justifications}
\label{app:justifications}
Verbatim free-text justifications from the Opus~5 controller, three per feedback variant, each the
text the model produced before its JSON edit (the JSON is omitted). Headers give assay, seed and
round.

\newcommand{\justif}[1]{\par\smallskip\noindent\textbf{#1}\par\nopagebreak}
\newenvironment{justquote}{\begin{quote}\small}{\end{quote}}

\paragraph{Full feedback (full LLM control).}

\justif{AACC1\_PSEAI\_Dandage\_2018, seed 0, round 1}
\begin{justquote}
With 96 labels and a high-dimensional PLM embedding, ridge at alpha=1.0 is badly under-regularized---the
huge MAEs (1.2--2.4) and near-zero within-stratum correlations look like variance, not a missing
feature. Since the scored metric is Spearman and the label distribution is clearly heavy-tailed at the
low-fitness end (quintiles 1--2 are near-random), I'm switching to rank-transformed targets with an
averaged ridge ensemble and much stronger regularization. The uncertainty estimate barely tracks error
(rho=0.18), so UCB/disagreement acquisition would be unreliable; instead I'll use diversity\_greedy to
spread the next 96 labels across embedding space, which is the right spend this early (1255 left in
pool, top-10\% recall already saturated). No pseudo-labels yet---self-training off a model this noisy
would just amplify its own bias.
\end{justquote}

\justif{A0A247D711\_LISMN\_Stadelmann\_2021, seed 1, round 1}
\begin{justquote}
With 96 labels against a high-dimensional PLM embedding, \texttt{alpha=1.0} is almost certainly
underregularized---the pooled CV Spearman of 0.54 with wide strata CIs looks like variance, not a data
problem, so I'd raise regularization sharply and average over an ensemble of ridges for stability.
Since the metric is rank-based and the errors concentrate in the compressed top quintiles (MAE
$\sim$0.6 with flat rank correlation there), fitting on rank-transformed targets should align the
objective with evaluation better than MSE on a skewed fitness distribution. I'm leaving curation alone:
dropping outliers or deduplicating by position would discard scarce labels, and no stratum is
significantly worse than pooled, so there's no systematic defect to patch yet. The uncertainty estimate
barely tracks error (Spearman 0.278), so UCB or ensemble-disagreement acquisition would be chasing
noise; with 1144 variants left and only 96 measured, this round is still about covering feature space,
so I'll use \texttt{diversity\_greedy} rather than exploiting a model I don't trust. Exploitation can
wait until the CV correlation is high enough to justify it.
\end{justquote}

\justif{A4GRB6\_PSEAI\_Chen\_2020, seed 2, round 1}
\begin{justquote}
\textbf{Justification.} With 96 labels and high-dimensional embeddings, \texttt{alpha=1.0} is almost
certainly under-regularized---the huge MAE ($\sim$1.5) and the flat/negative Spearman in the
middle-upper fitness quintiles look like variance, not bias, so I'm raising alpha sharply and switching
to \texttt{ensemble\_ridge} for a more stable fit. Since the metric is Spearman, fitting on
rank-transformed targets aligns the loss with evaluation and also blunts the influence of the extreme
low-fitness tail without discarding data. The uncertainty estimate is useless (Spearman(std, $|$error$|$)
= 0.009), so UCB and \texttt{ensemble\_disagreement} acquisition are off the table this round; with 3657
variants left and only 96 measured, coverage is worth more than exploitation, so
\texttt{diversity\_greedy} spreads the next batch across sequence/feature space---which should also help
the weak position-tercile-0 stratum and the poorly-sampled low-fitness region. I'm holding off on
pseudo-labels and outlier dropping until the base model is stable enough to trust.
\end{justquote}

\paragraph{No-menu feedback.}

\justif{A0A247D711\_LISMN\_Stadelmann\_2021, seed 0, round 1}
\begin{justquote}
With 96 labels against high-dimensional embeddings, ridge at alpha=1.0 is almost certainly
under-regularized---that fits the modest pooled Spearman plus the near-flat per-stratum profile (no
stratum is genuinely worse; all CIs overlap the pool), so the fix is global capacity, not targeted
curation. The ensemble uncertainty is uncalibrated (Spearman(std, $|$error$|$) = $-$0.029, and the
highest-std bin is not the highest-error bin), so any UCB/max-variance acquisition would be driven by
noise and I will not use it this round. The purchased labels are badly skewed toward the top of the
fitness range (quintile 4 holds 54 of 96, and that is exactly the worst-resolved stratum at 0.115), so
the batch is better spent on coverage than on exploitation---recall@top-10\% is already 1.0, meaning
there is nothing to gain by chasing the top now. I am therefore raising alpha to 20 and switching
acquisition to diversity-based exploration over the 1144 remaining variants to spread labels across
positions and mutant AA classes. Everything else stays fixed so the alpha change is cleanly
attributable.
\end{justquote}

\justif{A0A247D711\_LISMN\_Stadelmann\_2021, seed 1, round 1}
\begin{justquote}
\textbf{Justification.} With 96 labels and high-dimensional embedding features, the model is almost
certainly variance-limited, and the flat within-quintile Spearmans (0.16, $-$0.04 in the top two bins)
look like an over-fit head that only recovers the coarse dead-vs-functional split; raising ridge
regularization is the highest-expected-value single edit. The ensemble uncertainty tracks error only
weakly (rho=0.28, and bin 3's mean std overshoots its mean error), so an uncertainty/UCB acquisition
rule would be steering on noise. With 1144 variants left and many rounds to go, round 2 should still be
exploration-weighted, but structured: the random batch left coverage uneven (position tercile 1 and
charged/aromatic mutants are the weak strata), so a diversity/coverage-based batch over feature space
and position should improve the model more than either random or greedy exploitation. I'm holding off
on pseudo-labels and outlier dropping until the head is properly regularized, so the effect of this
edit is interpretable.
\end{justquote}

\justif{A4GRB6\_PSEAI\_Chen\_2020, seed 0, round 1}
\begin{justquote}
\textbf{Justification.} With 96 labels against a high-dimensional embedding+ZS feature vector, ridge at
alpha=1.0 is badly under-regularized---the tell is large MAE ($\sim$1.1--1.6) spread evenly across every
stratum, and near-zero within-quintile Spearman, i.e.\ the model only separates dead from functional and
fits noise elsewhere. No stratum is reliably worse than the pooled value (22\% of CIs excluding it is
about what noise alone gives), so there is no case for stratum-targeted curation or reweighting yet.
The ensemble uncertainty is uninformative (Spearman(std, $|$error$|$) = 0.08), so any UCB/max-variance
acquisition would be effectively random with extra bias---I am not using it. With only 96 of 3753
variants labeled and the model still coarse, this round should explore: diversity/coverage acquisition
over positions and substitution types gives a better-conditioned design matrix than random draws, and it
costs nothing in expectation versus random even if the model is weak. I am therefore raising alpha
substantially and switching acquisition to diversity, leaving curation and pseudo-labels untouched so
the regularization effect is measurable in isolation.
\end{justquote}

\paragraph{Scalar feedback (CV Spearman only, with menu).}

\justif{A0A247D711\_LISMN\_Stadelmann\_2021, seed 2, round 1}
\begin{justquote}
With only 96 labels against high-dimensional embeddings, the dominant failure mode is variance, not
bias---ridge at alpha=1.0 is almost certainly under-regularized, so I'm raising alpha sharply and using
an ensemble of ridges for stability. Since the metric is rank correlation and DMS assays have
compressed dead-variant tails plus heavy upper noise, fitting on rank-transformed targets with light
winsorization should align the objective with evaluation without discarding real signal. I'm leaving
the feature set at emb+zs: one-hot would add hundreds of sparse columns that 96 rows cannot support.
For acquisition, model-derived uncertainty from 96 points is not yet trustworthy enough to drive UCB or
disagreement sampling, and only $\sim$8\% of the pool is labeled, so this round is better spent on
coverage---diversity\_greedy to spread across sequence/position space before exploiting. I'll revisit
uncertainty-driven acquisition once $\sim$200+ labels make the ensemble spread meaningful.
\end{justquote}

\justif{A0A247D711\_LISMN\_Stadelmann\_2021, seed 0, round 1}
\begin{justquote}
With 96 labels against a high-dimensional PLM embedding, ridge at alpha=1.0 is badly
under-regularized---that's the most likely source of the mediocre 0.378, not data quality, so the
biggest single win is a much stronger penalty. Since the metric is Spearman, fitting rank-transformed
targets aligns the loss with evaluation and also absorbs the heavy lower tail typical of DMS assays,
which makes outlier dropping and winsorizing redundant for now. I'm switching to
\texttt{ensemble\_ridge} for variance reduction and to start building a usable disagreement signal for
later rounds, while keeping \texttt{emb+zs} (one-hot would add parameters we can't afford yet). For
acquisition, 96 labels are far too few for a trustworthy uncertainty estimate, and
\texttt{zero\_shot\_prior}/\texttt{greedy} would concentrate the batch on predicted-high variants and
leave the fitness range poorly covered; \texttt{diversity\_greedy} buys broad embedding coverage now,
and I'll move to \texttt{ucb}/\texttt{ensemble\_disagreement} once the ensemble is calibrated. No
pseudo-labels yet---adding 100s of zero-shot labels on top of 96 real ones would let the ZS prior
dominate before we know the head is sound.
\end{justquote}

\justif{A4GRB6\_PSEAI\_Chen\_2020, seed 1, round 1}
\begin{justquote}
With only 96 labels against a high-dimensional embedding, the dominant risk is variance, not bias:
ridge at alpha=1.0 is almost certainly under-regularized, so I'm raising alpha sharply and switching to
an averaged ridge ensemble. Since the metric is Spearman and DMS distributions have a heavy
dead-variant tail, fitting on rank-transformed targets aligns the loss with the objective and absorbs
outliers without discarding data. Uncertainty from a 96-point linear model is not trustworthy enough to
drive UCB or disagreement acquisition yet, and the model is still coverage-limited rather than
boundary-limited, so this batch should explore. diversity\_greedy spreads the next 96 measurements
across embedding space, which improves the design matrix for every later round; I'll revisit
uncertainty-driven acquisition once $\sim$300 labels make the variance estimates meaningful. No
pseudo-labels yet---self-training off a 0.57-Spearman model would mostly inject its own errors.
\end{justquote}

\end{document}